%% file: main.tex
\documentclass[10pt,twocolumn,letterpaper]{article}

\usepackage{cvpr}

\input{preamble}

\definecolor{cvprblue}{rgb}{0.21,0.49,0.74}
\usepackage[pagebackref,breaklinks,colorlinks,allcolors=cvprblue]{hyperref}

\def\paperID{*****}
\def\confName{CVPR}
\def\confYear{2026}

\title{Edit-VAR: Taming Visual Autoregressive Model for Precise Video Editing}

\author{
Chongbo Zhao$^{1,*}$ \quad
Jiangming Wang$^{1,*}$ \quad
Xilai Wang$^{2}$ \quad
Xinyu Wang$^{3}$ \\
Jingyi Tang$^{4}$ \quad
Chunjie Hao$^{5}$ \quad
Pengjie Song$^{6}$ \quad
Yue Ma$^{3,\dagger}$ \\[3pt]
$^{1}$Sun Yat-sen University \quad
$^{2}$South China University of Technology \\
$^{3}$Tsinghua University \quad
$^{4}$Shandong University \quad
$^{5}$Nankai University \\
$^{6}$Hunan University \\ \quad
$^{*}$Equal contribution. \quad $^{\dagger}$Corresponding author.
}

\begin{document}

\maketitle

\begin{strip}
\centering
\includegraphics[width=\linewidth]{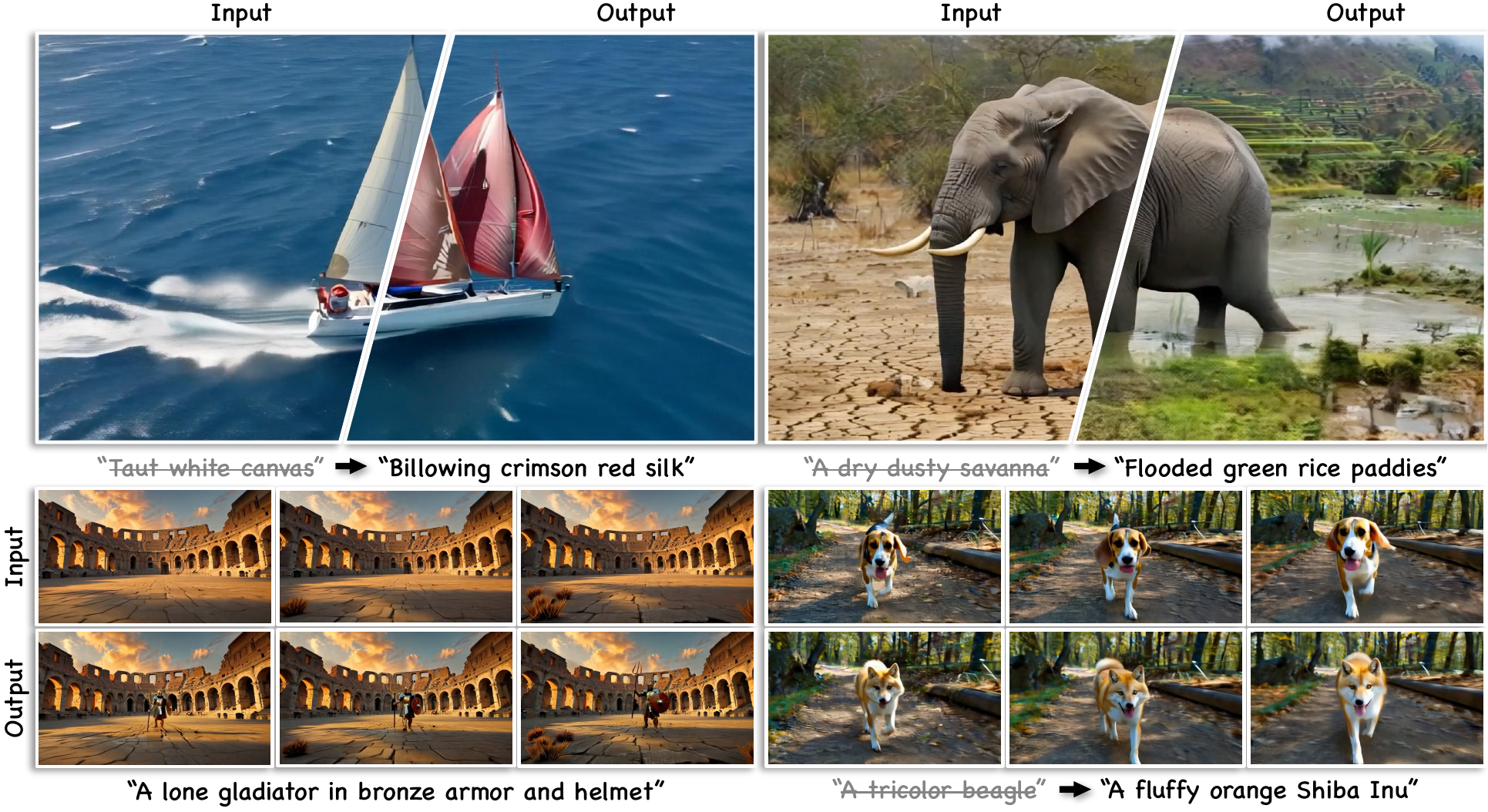}
\captionsetup{hypcap=false}
\captionof{figure}{Showcase of Edit-VAR. Our method supports attribute editing (top left), background replacement (top right), object addition (bottom left), and object replacement (bottom right), while preserving unedited content and temporal coherence.}
\captionsetup{hypcap=true}
\label{fig:qualitative_overview}
\end{strip}

\begin{abstract}
\input{sections/abstract}
\end{abstract}

\input{sections/introduction}

\input{sections/related_work}
\input{sections/method}
\input{sections/experiment}
\input{sections/conclusion}

{
    \small

\input{main.bbl}
}

\clearpage
\appendix
\section*{Supplementary Material}
\input{sections/supplementary}

\end{document}

%% file: preamble.tex
\PassOptionsToPackage{hyphens}{url}

\usepackage{graphicx}
\usepackage{amsmath}
\usepackage{amssymb}
\usepackage{booktabs}
\usepackage{multirow}
\usepackage{xcolor}
\usepackage{algorithm}
\usepackage{algorithmic}
\usepackage{placeins}

\usepackage{cuted}
\usepackage{caption}

\graphicspath{{images/}}

%% file: sections/abstract.tex
Text-guided video editing modifies target content while preserving the appearance and temporal coherence of unedited regions.
Training-based approaches provide strong control but demand substantial data and computation.
Training-free methods fall into inversion-free and inversion-based paradigms.
Inversion-free approaches avoid trajectory recovery, but their source-preserving guidance can limit editing strength and leave semantic changes incomplete.
Inversion-based approaches recover a latent trajectory before regeneration, where approximation errors can accumulate and cause source-content drift and temporal inconsistency.
We introduce \textbf{Edit-VAR}, the first training-free and inversion-free framework for text-guided video editing with a pretrained visual autoregressive video model.
Edit-VAR directly encodes the source video into multi-scale discrete tokens and performs probability-guided conditional token replacement for source preservation.
Attention-guided token-wise and scale-aware modulation selectively relaxes source constraints over edit-relevant positions and generation stages, while \emph{Scale-Decoupled Generation}, implemented as late-scale constraint release, regenerates motion-consistent details and reduces texture fragmentation.
Residual-guided token pruning further exploits redundancy at the final two high-resolution scales to reduce inference cost.
Extensive experiments and a blind user study demonstrate that Edit-VAR outperforms existing training-free video editing methods overall in editing fidelity, source preservation, temporal coherence, and inference efficiency.
Project page: \url{https://chongbozhao3-coder.github.io/Edit-VAR/}. Code: \url{https://github.com/chongbozhao3-coder/Edit-VAR}.

%% file: sections/introduction.tex
\section{Introduction}

Text-guided video editing modifies an existing video according to natural-language instructions, providing a flexible tool for content creation and post-production.
Its key challenge is to achieve the intended semantic change while preserving unrelated source content and temporal coherence.

Training-based approaches learn explicit editing behaviors from large-scale video data and generally provide strong instruction following and flexible control~\cite{ye2025unic,jiang2025vace,mai2025easyv2v}.
However, collecting such data and training dedicated video models require substantial computational resources.
Training-free methods instead reuse pretrained generative models and can be broadly divided into inversion-free and inversion-based paradigms.
Inversion-free approaches directly construct source-to-target transformations without recovering a noise trajectory~\cite{kulikov2025flowedit,li2025flowdirector}, but their source-preserving transport or guidance can restrict semantic deviation from the input, resulting in incomplete edits.
Inversion-based methods first map the source video onto a latent trajectory and then regenerate it under the target prompt~\cite{qi2023fatezero,tokenflow2023,kara2024rave,chen2026contextflow}; Approximation errors introduced during inversion and regeneration can propagate through the spatiotemporal representation, causing source-content drift and temporal inconsistency.
These limitations reveal a persistent trade-off between editability and source fidelity in existing training-free video editing methods.

\begin{figure}[t]
\centering
\includegraphics[width=\columnwidth]{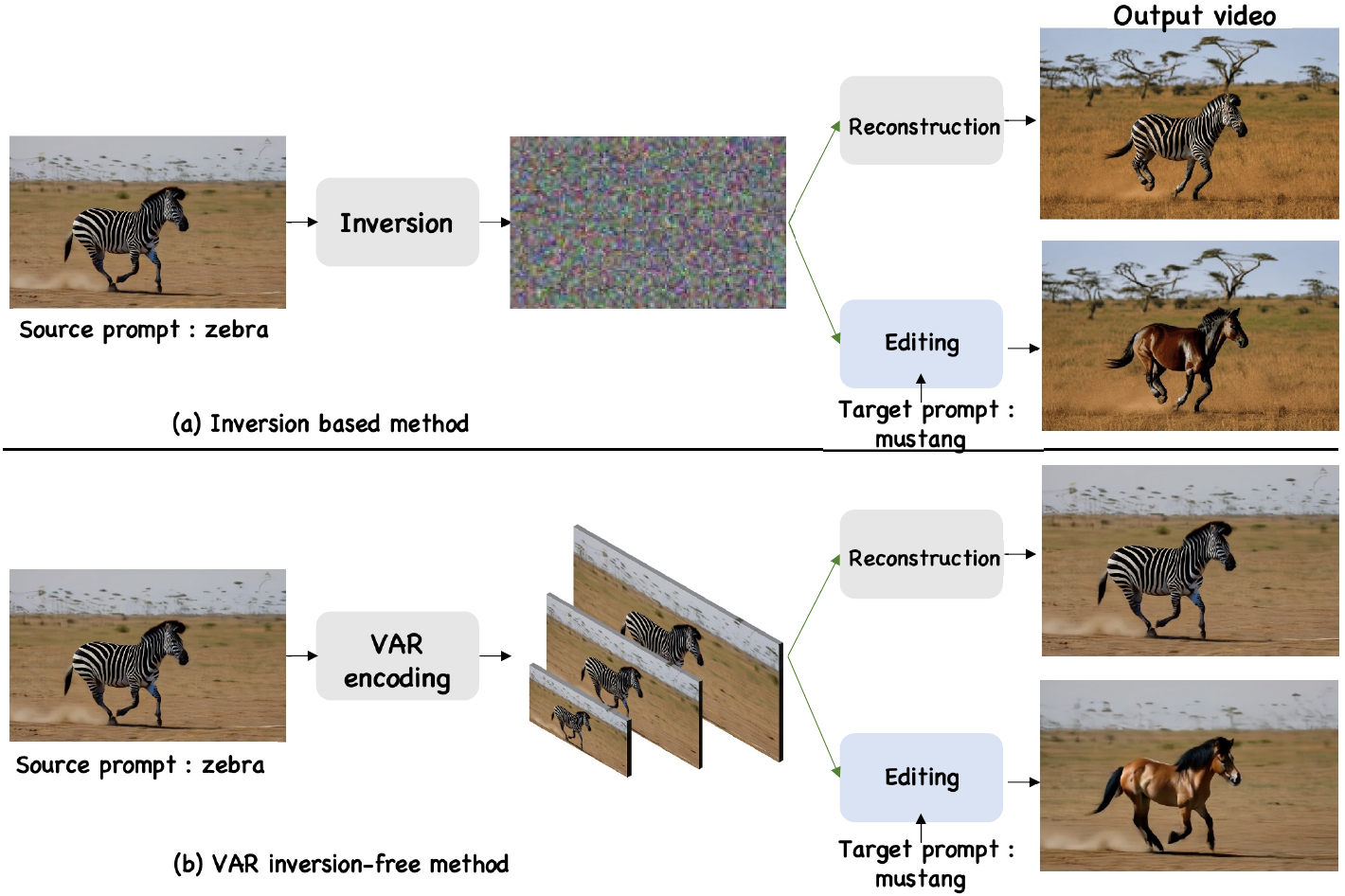}
\caption{Motivation. Comparison between continuous inversion-based editing and our discrete VAR formulation. Inversion and subsequent regeneration can introduce source-content drift, whereas direct VAR encoding provides multi-scale discrete source tokens for reconstruction and conditional replacement without iterative trajectory inversion.}
\label{fig:motivation}
\end{figure}

Figure~\ref{fig:motivation} motivates formulating video editing with a visual autoregressive (VAR) model~\cite{tian2024visual,han2025infinity,liu2026infinitystar}.
Its tokenizer directly encodes the source video into a coarse-to-fine hierarchy of discrete tokens, providing an explicit source reference without iterative trajectory inversion.
Editing can then be performed by selectively preserving or replacing these encoded tokens, avoiding inversion-induced reconstruction drift while retaining direct control over source fidelity.

Image-level AR editing~\cite{wang2025training} suggests probability-guided token replacement as a natural mechanism for training-free editing. However, directly extending this strategy to video introduces a different preservation problem. Video tokens are temporally coupled and jointly encode appearance, pose, motion, and edit regions that may evolve across frames. A uniform preservation rule can therefore over-constrain edit-relevant tokens, leaving semantic changes incomplete, or under-constrain non-edited tokens, causing background drift and temporal inconsistency. This suggests that VAR-based video editing requires preservation control that is both token-wise and scale-aware.

To bridge this gap, we introduce \textbf{Edit-VAR}, which reformulates probability-guided token replacement as attention-guided, token- and scale-adaptive preservation control over the spatiotemporal generation hierarchy, enabling coherent edits across frames while retaining unrelated source content.
Edit-VAR uses source-to-edit probability changes to assess token compatibility and perform conditional token replacement.
Cross-attention maps and a scale-aware envelope instantiate this control by modulating preservation strength across token positions and generation stages.

A further challenge arises at later high-resolution scales, where source tokens encode fine textures tightly coupled to the original motion and pose.
Forcing these tokens into the edited content can cause texture fragmentation across frames.
We therefore introduce Scale-Decoupled Generation, which releases source constraints beyond a designated scale and regenerates motion-consistent target details.
Finally, because the highest-resolution scales dominate inference cost while many tokens require little refinement, residual-guided token pruning concentrates computation on active positions to improve inference efficiency.

Our contributions are as follows:
\begin{itemize}
    \item We introduce \textbf{Edit-VAR}, the first training-free, inversion-free framework for text-guided editing with a pretrained VAR video model. It reformulates editing as conditional replacement of directly encoded source tokens, avoiding iterative trajectory inversion.
    \item We develop hierarchical adaptive preservation, combining probability-guided replacement with attention-guided token- and scale-aware control for localized and temporally coherent editing. Scale-Decoupled Generation regenerates motion-consistent details, while residual-guided token pruning removes redundant late-scale computation.
    \item Experiments across four editing tasks and a blind user study show that Edit-VAR achieves the best target alignment and non-edit preservation among the compared methods, competitive video quality, and a $2.04\times$ end-to-end speedup.
\end{itemize}

%% file: sections/related_work.tex
\section{Related Work}

\textbf{Image and Video Editing.}
Existing image and video editing methods span training-based, source-specific optimization, and training-free paradigms.
Training-based methods fine-tune pretrained generative models or train dedicated editing networks on paired data
\cite{ye2025unic,mai2025easyv2v,ju2025editverse,jiang2025vace,ma2024followpose,ma2025followcreation,ma2026fastvmt,ma2025followyourmotion,ma2025controllable},
but require substantial training data and task-specific optimization.
Other approaches adapt pretrained generators to each input through source-specific tuning or inversion
\cite{liu2024video,feng2025dit4edit,wang2024taming}.
Training-free methods avoid dataset-level retraining and control editing through attention manipulation or feature reuse
\cite{hertz2022prompt,cao2023masactrl,ma2026livelight,ma2025followfaster,wang2024cove,yang2025unified,wang2026liveedit,liu2026opsd,qi2023fatezero,tokenflow2023,chen2026contextflow,zhu2025kv}.
Most existing approaches recover a source trajectory through inversion before regeneration, where numerical approximation may introduce unintended changes in non-target regions.
Recent inversion-free methods bypass trajectory recovery
\cite{kulikov2025flowedit,li2025flowdirector,zhu2025kv,qiu2024tfb,qiu2025duet,qiu2025DBLoss,qiu2026dag},
but still face a preservation--editability trade-off: strong source constraints may suppress the requested edit, whereas weaker constraints can degrade source fidelity.
These limitations motivate inversion-free editing with explicit source-preservation control.
\paragraph{Visual AutoRegressive Generation and Editing.}
Visual autoregressive models represent images as discrete token sequences and generate them through next-token or next-scale prediction.
The visual autoregressive (VAR) paradigm, pioneered by VAR~\cite{tian2024visual} and extended by Infinity~\cite{han2025infinity}, redefines generation as a coarse-to-fine multi-scale process. InfinityStar~\cite{liu2026infinitystar} further unifies this framework across the spatiotemporal dimension.
The discrete token formulation naturally supports direct token replacement without reconstructing a continuous generative trajectory.
With the maturation of AR generation models, leveraging them for image editing has become a natural research direction. Training-based methods such as ControlVAR~\cite{li2024controlvar} incorporate conditional controls but depend on paired training data.
AREdit~\cite{wang2025training} enables training-free image editing by caching source-token probabilities and deriving adaptive token-replacement decisions, while ISLock~\cite{hu2025anchor} preserves source structure through anchor-token matching. However, these image-oriented formulations do not explicitly account for the spatially varying edit regions and scale-dependent source constraints encountered in video generation. Directly applying such preservation strategies to video may either insufficiently localize the intended edit or over-constrain fine-scale details that are coupled with source motion and pose. Our method addresses these limitations through attention-guided token-wise modulation and scale-decoupled generation.
Visual AR acceleration reduces inference redundancy through token or layer pruning and cache compression~\cite{guo2025fastvar,song2026vista,song2026streamingeffect,gao2026pai,song2024processpainter,yune2026faststar,chen2026toprovar,qin2026head}. Our method instead uses residual norms to select tokens for pruning at the final high-resolution scales.

%% file: sections/method.tex
\section{Method}

\begin{figure*}[t]
\centering
\includegraphics[width=\textwidth]{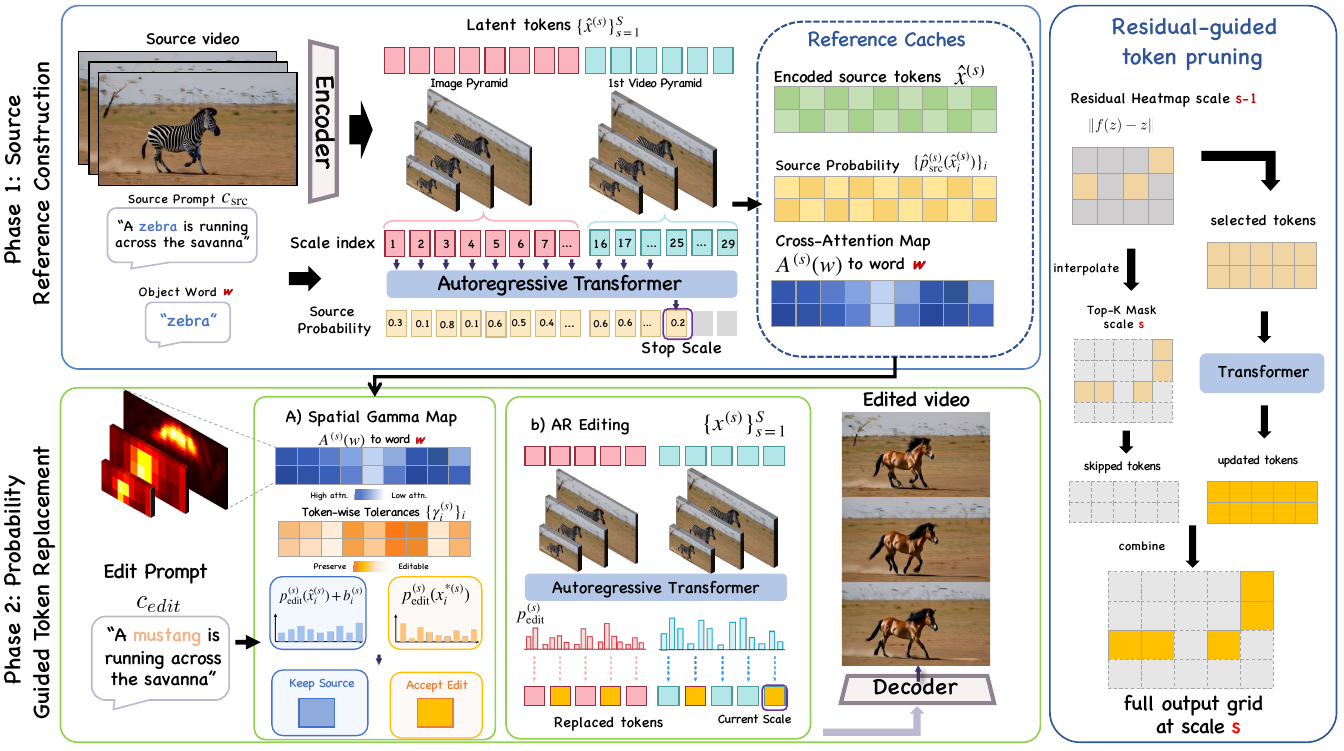}
\caption{Overview of our framework. In Phase~1, the source video is encoded and a forward pass under the source prompt caches per-token probabilities and cross-attention maps. In Phase~2, a forward pass under the edit prompt performs probability-guided token replacement steered by the attention-guided spatial modulation field, which assigns low $\gamma$ to edited regions and high $\gamma$ to preserved regions. Source caching stops at scale $S_{\mathrm{stop}}$; beyond that, the editing pass generates freely.}
\label{fig:framework}
\end{figure*}

Our method edits a source video directly in the discrete token space of a visual autoregressive model without any additional training.

\subsection{Preliminary}
\label{sec:preliminary}

Our method builds on InfinityStar~\cite{liu2026infinitystar}, a visual autoregressive video model that generates videos as a sequence of $S$ discrete token blocks $\mathbf{x} = \{x^{(s)}\}_{s=1}^{S}$ from coarse to fine:
\begin{equation}
p(\mathbf{x} \mid c) = \prod_{s=1}^{S} p(x^{(s)} \mid x^{(<s)}, c),
\label{eq:ar}
\end{equation}
where $c$ is a text prompt and all tokens within each scale are predicted in parallel.
Early scales encode global structure while later high-resolution scales refine fine details.
The model uses a spacetime pyramid with an image tower (scales $1$ to $S_{\mathrm{img}}$, single frame) followed by a video tower (scales $S_{\mathrm{img}}\!+\!1$ to $S$, multi-frame), each with increasing spatial resolution.

Two properties make this architecture particularly suited for editing.
First, the tokenizer's encoder directly maps any video to deterministic source token indices at every scale---no iterative trajectory inversion is needed to recover the source representation.
Second, at each generation step tokens are selected by $\arg\max$ over a probability distribution; our method intervenes at this selection step to steer preservation or replacement.

\subsection{Probability-Guided Video Editing}
\label{sec:editing}

Given a source video $v$, a source prompt $c_{\mathrm{src}}$, and an edit prompt $c_{\mathrm{edit}}$, our method proceeds in two forward passes.
Phase~1 encodes $v$ into encoded source tokens and caches the model's source-conditioned probabilities and cross-attention maps.
Phase~2 runs under the edit prompt, using a probability-guided preservation mechanism to decide per-token preservation or replacement.

\paragraph{Source Reference Construction (Phase~1).}
The tokenizer's encoder maps $v$ to encoded source tokens indices $\hat{x}^{(s)}$ at every scale.
We then run a forward pass under $c_{\mathrm{src}}$, force-decoding with $\hat{x}^{(s)}$ at each scale to obtain the source-conditioned probability $\hat{p}^{(s)}_{\mathrm{src}} = p_\theta(x^{(s)} \mid \hat{x}^{(<s)}, c_{\mathrm{src}})$.
During this pass, we also extract cross-attention maps $A^{(s)}$ for a source-side attention anchor, which localize the editing region in the video.
We obtain the anchor automatically by segmenting the source and edit prompts into words and computing their word-level difference. For replacement, attribute, and background edits, the source-side word or phrase in the differing span is used as the anchor. For object addition, where the inserted span has no source-side counterpart, the algorithm uses the nearest unchanged source-side content word adjacent to the insertion span. This procedure is deterministic and requires neither manual anchor selection nor an additional language model.
For scales whose sequence length exceeds the attention window, attention maps are interpolated from the nearest available scale.
The cached triplet $\{\hat{x}^{(s)},\, \hat{p}^{(s)}_{\mathrm{src}},\, A^{(s)}\}$ serves as the source reference for Phase~2.

\paragraph{Probability-Guided Preservation.}
The preservation decision is based on the source-to-edit support drop of the cached source token.
If the edit prompt assigns similar support to a cached source token as the source prompt, the token is likely still compatible with the intended edit and should remain competitive for preservation.
Conversely, a substantial support drop indicates that the edit prompt no longer supports the source token, suggesting that it should be more open to replacement.

We formalize this drop-based compatibility test using a local preservation tolerance $\gamma$.
At each position $i$ in scale $s$, we compute a bias toward the source token and compare it against the edit prediction $x^{*(s)}_i = \arg\max_j\, p^{(s)}_{\mathrm{edit}}(j)$:
\begin{align}
b^{(s)}_i &= \max\!\bigl(\gamma^{(s)}_i - \hat{p}^{(s)}_{\mathrm{src}}(\hat{x}^{(s)}_i),\; 0\bigr), \label{eq:bias} \\
x^{(s)}_i &= \begin{cases} \hat{x}^{(s)}_i & \text{if } p^{(s)}_{\mathrm{edit}}(\hat{x}^{(s)}_i) + b^{(s)}_i \geq p^{(s)}_{\mathrm{edit}}(x^{*(s)}_i), \\[3pt] x^{*(s)}_i & \text{otherwise.} \end{cases} \label{eq:select}
\end{align}
When the bias is positive, the adjusted score of the source token can be rewritten as
\begin{equation}
p^{(s)}_{\mathrm{edit}}(\hat{x}^{(s)}_i) + b^{(s)}_i
= \gamma^{(s)}_i - \Bigl[\hat{p}^{(s)}_{\mathrm{src}}(\hat{x}^{(s)}_i) - p^{(s)}_{\mathrm{edit}}(\hat{x}^{(s)}_i)\Bigr].
\label{eq:prob_drop}
\end{equation}
Eq.~\eqref{eq:prob_drop} shows that the source token is not selected according to its absolute probability alone, but according to how much its support decreases from the source prompt to the edit prompt.
A larger source-to-edit support drop reduces the adjusted competitiveness of the cached source token, making replacement more likely.
The tolerance $\gamma^{(s)}_i$ controls how much drop is allowed before source preservation is relaxed: larger values favor preservation under larger drops, while smaller values make the token more sensitive to edit-induced support changes.
When $\gamma^{(s)}_i = 0$, no preservation bias is applied and the edit distribution decides freely.
When $\gamma^{(s)}_i = 2.0$, preservation is enforced because all probabilities are bounded by one.
\paragraph{Attention-Guided Spatial Modulation.}
A single $\gamma$ for all tokens cannot distinguish edited regions from preserved regions.
We observe that cross-attention maps associated with the source-side edit target naturally highlight the spatial positions associated with the editing target---tokens with stronger responses are more likely to belong to the edit-relevant region, low-attention tokens generally correspond to source content that should be preserved.
This provides a training-free spatial editing signal without requiring any external segmentation.
We leverage this by mapping attention scores to per-token gamma values: high attention (editing target) $\to$ low $\gamma$ (allow editing), low attention (background) $\to$ high $\gamma$ (preserve).
This is implemented as a two-level sigmoid.

\noindent\textit{Level 1: Scale Envelope.}\quad
Each scale $s$ has a gamma range $[\gamma^{(s)}_{\mathrm{low}},\, \gamma^{(s)}_{\mathrm{high}}]$ determined by scale position:
\begin{equation}
\gamma_{\{\mathrm{low},\mathrm{high}\}}^{(s)} = \gamma_{\mathrm{start}} + \bigl(\gamma_{\mathrm{end}}^{\{\mathrm{fg},\mathrm{bg}\}} - \gamma_{\mathrm{start}}\bigr) \cdot \sigma\!\Bigl(\frac{t - t_c}{w_t}\Bigr),
\label{eq:scale_envelope}
\end{equation}
where $s_{\mathrm{local}} \in \{0,\ldots,N_{\mathrm{tower}}-1\}$ is the local scale index within the current image or video tower, $N_{\mathrm{tower}}$ is the number of scales in that tower, and $t = s_{\mathrm{local}}/(N_{\mathrm{tower}}\!-\!1)$ is the normalized scale position. We set $\gamma_{\mathrm{start}} = 2.0$ so that early scales are fully preserved.
At later high-resolution scales, $\gamma^{(s)}_{\mathrm{low}}$ drops to allow editing while $\gamma^{(s)}_{\mathrm{high}}$ stays high for background preservation.

\noindent\textit{Level 2: Token-wise Mapping.}\quad
Within each scale's range, individual tokens are mapped by their attention score $a_i \in [0,1]$:
\begin{equation}
\gamma^{(s)}_i = \gamma^{(s)}_{\mathrm{high}} + (\gamma^{(s)}_{\mathrm{low}} - \gamma^{(s)}_{\mathrm{high}}) \cdot \sigma\!\Bigl(\frac{a_i - c_a}{w_a}\Bigr).
\label{eq:token_gamma}
\end{equation}
\paragraph{Scale-Decoupled Generation.}
Source tokens at later high-resolution scales encode fine textures tightly coupled to the source video's motion.
Forcing their preservation when the edited content has different dynamics causes texture fragmentation.
We stop Phase~1 at scale $S_{\mathrm{stop}}$: beyond this point, no source probabilities or attention are cached, and Phase~2 generates freely ($\gamma = 0$).
The source constraints applied at early scales largely preserve the global layout, while fine details are regenerated to match the edited content.

\begin{figure}[t]
\centering
\includegraphics[width=\columnwidth]{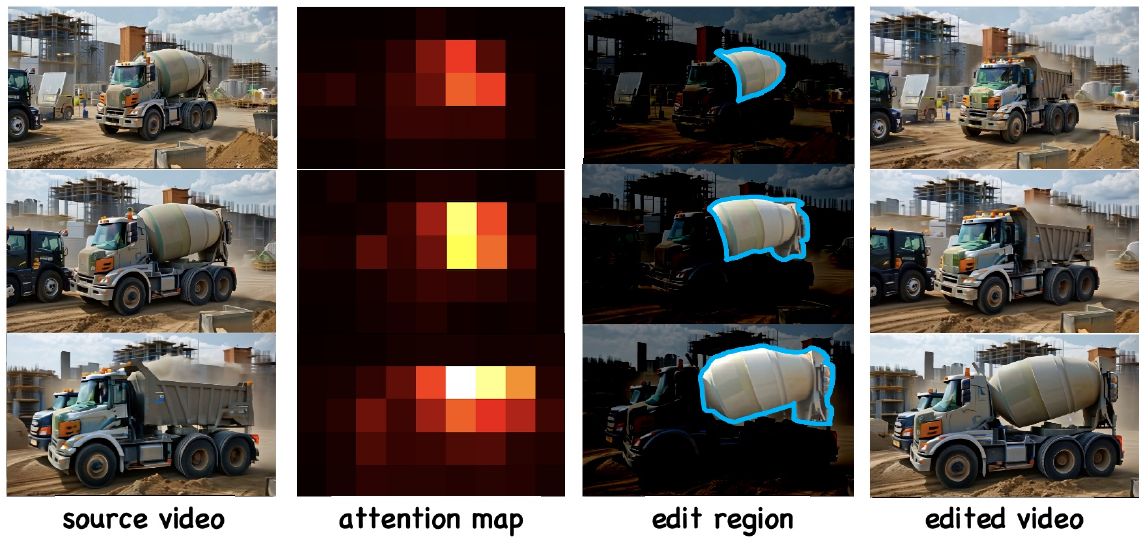}
\caption{Spatial modulation visualization. Given a source frame and an edit-target word, the cross-attention map localizes the editing region, which is then mapped to a per-token gamma field.}
\label{fig:spatial_modulation}
\end{figure}

Algorithms~1 and~2 in the supplementary material summarize the complete procedure.

\input{sections/experiment_floats}

\subsection{Inference Acceleration}
\label{sec:acceleration}

Beyond Scale-Decoupled Generation, which reduces Phase~1 runtime from 65.0\,s to 19.8\,s ($3.29\times$) by skipping transformer computation at scales $s \geq S_{\mathrm{stop}}$, we analyze computational redundancy in the editing pass and introduce an acceleration strategy for later high-resolution scales.

\paragraph{Residual-Guided Token Pruning.}
At later high-resolution scales, not all spatial positions require equal computation---positions where the transformer makes substantial changes are ``active,'' while positions with small residuals require less refinement.
We use the transformer's own residual as a direct signal of computational necessity.
For each of the final two generation scales $s$, we use the preceding scale's per-token residual norm $\|f_\theta(x)-x\|$ as the activity signal.
This residual map is trilinearly interpolated to scale $s$'s spatial dimensions, and only the top-$K\%$ tokens by residual magnitude are forwarded through the transformer; the remaining tokens retain their input embeddings unchanged.
Keep indices are fixed across all 36 transformer blocks. Pruned tokens bypass current-scale Q/K/V, attention, and feed-forward computation, and their input states are passed directly to the logits head; retained tokens still access the cross-scale KV cache from earlier scales.
We use $K=50\%$ by default, thereby skipping half of the tokens at these two highest-resolution scales.
This preserves the full 2D/3D token lattice structure while concentrating computation on active regions and reducing both editing-pass and end-to-end inference time.

%% file: sections/experiment_floats.tex
\begin{table*}[t]
    \centering
    \setlength{\tabcolsep}{4.2pt}
    \begin{tabular}{l cc cc ccc c}
    \toprule
    \multirow{2}{*}{Method} & \multicolumn{2}{c}{Target Alignment} & \multicolumn{2}{c}{Video Quality} & \multicolumn{3}{c}{Non-edit Preservation} & \multirow{2}{*}{Time$\downarrow$} \\
    \cmidrule(lr){2-3} \cmidrule(lr){4-5} \cmidrule(lr){6-8}
    & CLIP-S$_{\mathrm{full}}\uparrow$ & CLIP-S$_{\mathrm{edit}}\uparrow$ & Smooth.$\uparrow$ & Aesth.$\uparrow$ & PSNR$\uparrow$ & SSIM$\uparrow$ & LPIPS$\downarrow$ & \\
    \midrule
    \multicolumn{9}{l}{\textit{Training-based reference}} \\
    VACE-V2V     & 0.957 & 0.970 & 0.936 & 0.605 & 13.51 & 0.380 & 0.618 & 9 min \\
    \midrule
    \multicolumn{9}{l}{\textit{Training-free, inversion-based}} \\
    RAVE         & 0.962 & 0.975 & 0.973  & 0.552 & 18.81 & 0.674 & 0.437 & 17 min \\
    FADE         & 0.966 & 0.972 & 0.958 & 0.533 & 14.75 & 0.626 & 0.211 & 14 min \\
    \midrule
    \multicolumn{9}{l}{\textit{Training-free, inversion-free}} \\
    Wan-Edit     & 0.970 & 0.966 & 0.988 & 0.607 & 21.04 & 0.801 & 0.183 & 183 s \\
    FlowDirector & 0.965 & 0.967 & 0.983 & 0.594 & 21.74 & 0.814 & 0.192 & 37 min \\
    Ours & 0.972 & 0.976 & 0.988 & 0.609 & 22.58 & 0.821 & 0.176 & 64 s \\
    \bottomrule
    \end{tabular}
    \caption{Main quantitative comparison with wall-clock inference time.}
    \label{tab:main}
    \end{table*}
    
    \begin{figure*}[t]
    \centering
    \includegraphics[width=\textwidth]{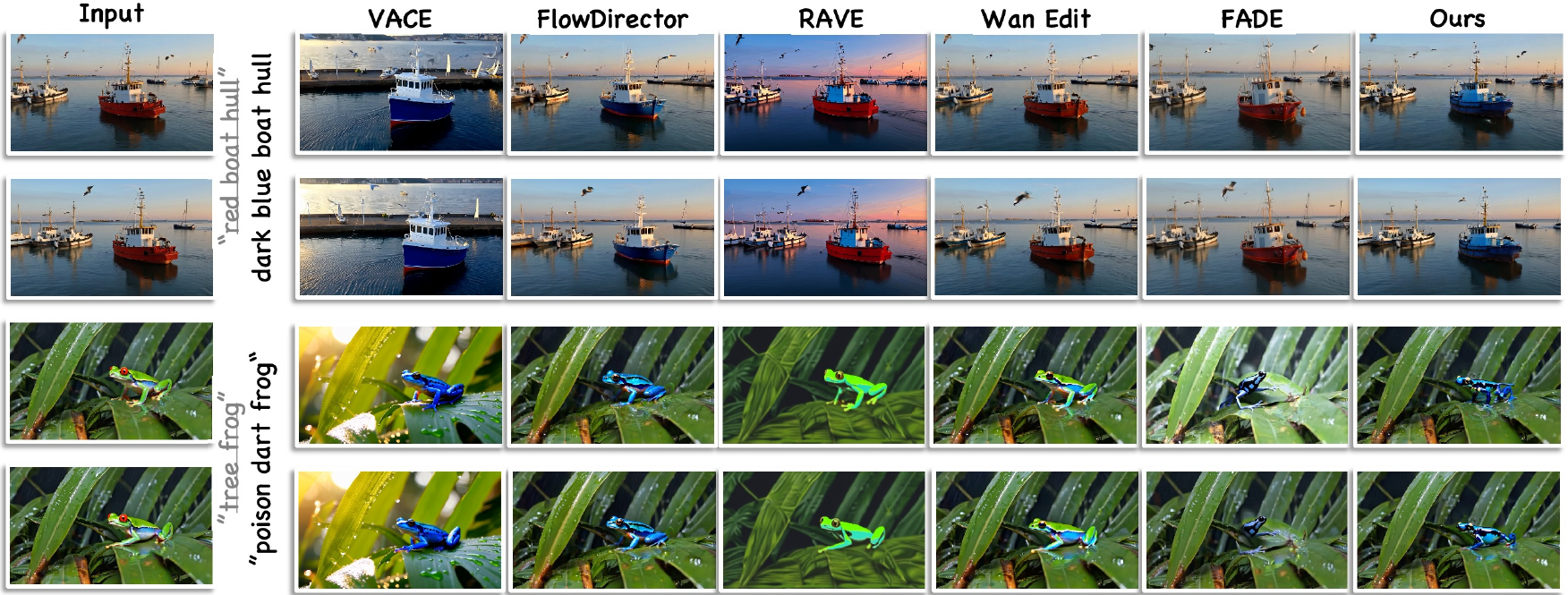}
    \caption{Qualitative comparison. Each example shows two frames for the edits ``red boat hull'' $\rightarrow$ ``dark blue boat hull'' and ``tree frog'' $\rightarrow$ ``poison dart frog.'' Edit-VAR consistently realizes the target appearance across frames while preserving scene geometry and unrelated background content. Competing methods may leave source attributes incompletely edited or introduce unintended changes to the background, lighting, and object appearance.}
    \label{fig:comparison}
    \end{figure*}
    
    \begin{table*}[t]
    \centering
    \setlength{\tabcolsep}{1pt}
    \begin{tabular}{l cc cc ccc c}
    \toprule
    \multirow{2}{*}{Variant} & \multicolumn{2}{c}{Target Alignment} & \multicolumn{2}{c}{Video Quality} & \multicolumn{3}{c}{Non-edit Preservation} & \multirow{2}{*}{Time (s)} \\
    \cmidrule(lr){2-3} \cmidrule(lr){4-5} \cmidrule(lr){6-8}
    & CLIP-S$_{\mathrm{full}}\uparrow$ & CLIP-S$_{\mathrm{edit}}\uparrow$ & Smooth.$\uparrow$ & Aesth.$\uparrow$ & PSNR$\uparrow$ & SSIM$\uparrow$ & LPIPS$\downarrow$ & \\
    \midrule
    Full method (no pruning)           & 0.979 & 0.980 & 0.989 & 0.611 & 22.89 & 0.833 & 0.172 & 85s \\
    w/o attention guidance             & 0.935 & 0.931 & 0.975 & 0.560 & 19.56 & 0.563 & 0.373 & 85s \\
    w/o scale envelope                 & 0.958 & 0.933 & 0.979 & 0.514 & 20.43 & 0.667 & 0.287 & 85s \\
    w/o Scale-Decoupled Generation     & 0.935 & 0.951 & 0.980 & 0.524 & 21.43 & 0.612 & 0.189 & 130s \\
    AREdit-style uniform $\gamma$      & 0.895 & 0.913 & 0.975 & 0.541 & 16.30 & 0.498 & 0.456 & 85s \\
    \midrule
    + Random pruning ($K=50\%$)        & 0.943 & 0.954 & 0.976 & 0.589 & 17.73 & 0.781 & 0.235 & 64s \\
    + Residual pruning ($K=70\%$)      & 0.975 & 0.977 & 0.989 & 0.608 & 22.70 & 0.825 & 0.175 & 71s \\
    + Residual pruning ($K=50\%$, ours) & 0.972 & 0.976 & 0.988 & 0.609 & 22.58 & 0.821 & 0.176& 64s \\
    \bottomrule
    \end{tabular}
    \caption{Ablation study. The uniform-$\gamma$ variant is the AREdit-style baseline. $K$ denotes the percentage of retained tokens; random pruning controls for the token budget.}
    \label{tab:ablation}
    \end{table*}
    
    \begin{figure*}[t]
    \centering
    \includegraphics[width=\textwidth]{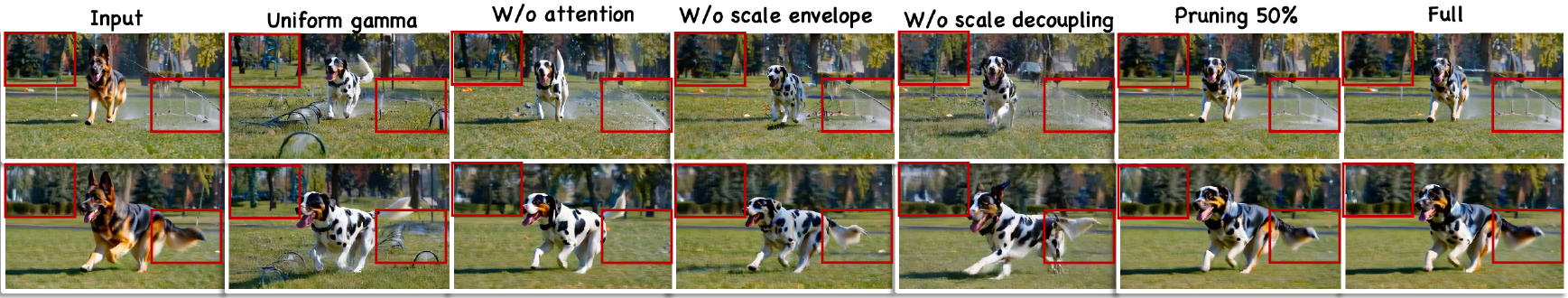}
    \caption{Qualitative ablation. Attention guidance and the scale envelope enable localized token replacement, while Scale-Decoupled Generation reduces fragmented source appearance. Residual pruning with $K=50\%$ remains visually close to the full method.}
    \label{fig:ablation_editing}
    \end{figure*}

%% file: sections/experiment.tex

\section{Experiments}

\subsection{Experimental Setup}
\label{sec:setup}

\paragraph{Implementation Details.}
Our method builds on InfinityStar~\cite{liu2026infinitystar}, a visual autoregressive video generation model, and generates 81-frame videos.
The key parameters are: $\gamma_{\mathrm{start}} = 2.0$, $\gamma_{\mathrm{end}}^{\mathrm{fg}} = 1.60$, $\gamma_{\mathrm{end}}^{\mathrm{bg}} = 1.78$, with sigmoid-based scale-envelope transitions at scale 5 (image tower) and scale 4 (video tower).
For Scale-Decoupled Generation, we set the decoupling boundary to $S_{\mathrm{stop}} = 25$, such that source caching is disabled for $s\geq 25$.
Residual-guided token pruning is applied only at the final two generation scales, retaining the top $K=50\%$ tokens by residual magnitude at each scale.
All parameters are fixed across all editing cases.
Experiments are conducted on a single NVIDIA A800 GPU.

\paragraph{Evaluation Dataset and Baselines.}
We collect 160 source videos from publicly accessible online sources. The evaluation covers object replacement, object addition, background replacement, and attribute editing, with 40 cases in each category.
We compare with five representative baselines under the same user-input setting, using only the source video and textual conditions. The training-based baseline is VACE-V2V~\cite{jiang2025vace}; the training-free inversion-based baselines are RAVE~\cite{kara2024rave} and FADE~\cite{zhu2025fadefrequencyawarediffusionmodel}; and the training-free inversion-free baselines are Wan-Edit~\cite{li2025five} and FlowDirector~\cite{li2025flowdirector}.
All baselines are evaluated using their official implementations and author-recommended configurations without case-specific tuning.

\paragraph{Evaluation Metrics.}
We evaluate target-prompt alignment using both full-frame and edit-region CLIP-Score (CLIP-S)~\cite{radford2021learning}, following the region-aware protocol of FiVE-Bench~\cite{li2025five}. The former measures global agreement with the complete target prompt, while the latter focuses on the intended edit. We assess temporal and perceptual video quality using Motion Smoothness and Aesthetic Quality from VBench~\cite{huang2024vbench}. Editing-region masks are generated once for each case using SAM2~\cite{ravi2024sam2} and shared across all methods. PSNR, SSIM~\cite{1284395}, and LPIPS~\cite{zhang2018unreasonable} are computed over the complementary non-edited regions to measure source-content preservation. All outputs use the same temporal sampling, masks, and metric implementations. Further details of mask construction and metric aggregation are provided in the supplementary material.

\subsection{Comparison with Baselines}
\label{sec:comparison}

Table~\ref{tab:main} shows that our final accelerated configuration achieves the strongest overall performance among the compared training-free methods. Edit-VAR obtains the highest full-frame and edit-region CLIP-S scores (0.972 and 0.976), indicating accurate realization of the target edit. It also achieves the highest aesthetic quality (0.609) and ties for the best motion smoothness (0.988). For non-edit preservation, Edit-VAR consistently ranks first across PSNR, SSIM, and LPIPS, reaching 22.58, 0.821, and 0.176, respectively. Meanwhile, it completes an 81-frame video edit in only 64 s, compared with 5–37 min for the training-free baselines. 

Figure~\ref{fig:comparison} further shows that Edit-VAR more reliably confines changes to the intended region: inversion-based methods often introduce background drift, whereas inversion-free baselines may leave the requested semantic change incomplete.

\paragraph{User Study.}
We conduct a blind user study with 15 participants on 50 randomly selected cases. Participants view the source video, target prompt, and an anonymized edited result at a fixed size and frame rate, and rate each result independently as \emph{Fail}, \emph{Pass}, \emph{Good}, or \emph{Excellent} in editing quality, background preservation, and temporal consistency. Our method receives the largest combined proportion of \emph{Good} and \emph{Excellent} ratings across all three criteria; complete results and protocols are provided in the supplementary material.

\subsection{Ablation Study}
\label{sec:ablation_study}

Table~\ref{tab:ablation} and Figure~\ref{fig:ablation_editing} evaluate both the editing and acceleration components of Edit-VAR.

\paragraph{Editing Components.}
Removing attention guidance degrades edit-region CLIP-S from 0.980 to 0.931 and SSIM from 0.833 to 0.563, confirming the value of explicit localization. Removing the scale envelope likewise reduces edit-region CLIP-S by 0.047 and SSIM by 0.166. The AREdit-style uniform-$\gamma$ baseline directly extends its spatially uniform preservation policy to video tokens; it gives the lowest edit-region CLIP-S (0.913) and highest LPIPS (0.456), showing the need to adapt preservation strength across edited and preserved regions. These results establish the complementarity of token-wise modulation and scale-aware preservation.

\paragraph{Scale-Decoupled Generation.}
Releasing source constraints at $S_{\mathrm{stop}}=25$ improves full-frame/edit-region CLIP-S from 0.935/0.951 to 0.979/0.980 and increases SSIM from 0.612 to 0.833. It also reduces Phase~1 runtime from 65.0\,s to 19.8\,s ($3.29\times$), lowering the two-phase runtime from 130.0\,s to 84.8\,s ($1.53\times$). Figure~\ref{fig:ablation_editing} shows the corresponding reduction in texture fragmentation.

\paragraph{Residual-Guided Token Pruning.}
Many tokens at the final high-resolution scales have near-zero transformer residuals. Retaining 70\% and 50\% of tokens accelerates Phase~2 by $1.27\times$ and $1.48\times$, reducing it from 65.0\,s to 51.2\,s and 43.9\,s, respectively. With Scale-Decoupled Generation applied in Phase~1, the corresponding two-phase runtimes are 70.9\,s and 63.7\,s, or $1.19\times$ and $1.33\times$ faster than the unpruned method. At the same 50\% token budget, random pruning substantially degrades edit-region CLIP-S (0.954 vs.~0.976) and LPIPS (0.235 vs.~0.176), supporting residual-based selection. We use $K=50\%$ as the default operating point.

\paragraph{Overall Efficiency.}
Table~\ref{tab:main} reports wall-clock time on the same hardware. Our final accelerated configuration requires 64\,s, compared with 5--37\,min for the baselines; the corresponding unpruned configuration requires 85\,s. Relative to the fully unoptimized two-phase runtime of 130\,s, combining Scale-Decoupled Generation and 50\% token pruning gives an overall $2.04\times$ speedup.

%% file: sections/conclusion.tex

\section{Conclusion}

We presented Edit-VAR, a training-free video editing framework that operates on directly encoded tokens of a visual autoregressive model. Attention-guided and scale-aware preservation balances editability with source fidelity, while late-scale constraint release reduces texture fragmentation and residual-guided pruning improves efficiency. Edit-VAR achieves the strongest average target alignment and non-edit preservation among the compared methods. These results demonstrate the potential of multi-scale discrete source tokens as an effective representation for training-free video editing.
\paragraph{Limitations.}
Our method formulates editing as conditional token replacement, which preserves the global structure of the source video by design.
This makes it unsuitable for edits that require large structural changes, such as object removal or significant layout modification, where the vacated region must be coherently inpainted rather than replaced token-by-token.
Extending the framework to support such structural edits is a promising direction for future work.

%% file: sections/supplementary.tex

\section{Algorithm Pseudocode}
\label{sec:supp_algorithm}
Algorithms~\ref{alg:phase1}--\ref{alg:pruning} summarize source-reference construction, probability-guided token replacement, and residual-guided token pruning.

\begin{algorithm}[ht]
\caption{Phase 1: Source Reference Construction}
\label{alg:phase1}
\begin{algorithmic}[1]
\REQUIRE Source video $v$; source prompt $c_{\mathrm{src}}$
\REQUIRE Edit prompt $c_{\mathrm{edit}}$; boundary $S_{\mathrm{stop}}$
\ENSURE Cached tokens and source probabilities
\ENSURE Cross-attention maps $\{A^{(s)}\}$
\STATE $\{\hat{x}^{(s)}\}_{s=1}^{S} \leftarrow \mathrm{Encode}(v)$
\STATE $w \leftarrow \mathrm{AutoAnchor}(c_{\mathrm{src}},c_{\mathrm{edit}})$
\FOR{$s = 1$ to $S_{\mathrm{stop}}-1$}
    \STATE Force-decode $\hat{x}^{(s)}$ under $c_{\mathrm{src}}$
    \STATE Cache $\hat{p}^{(s)}_{\mathrm{src}}$ and $A^{(s)}(w)$
\ENDFOR
\end{algorithmic}
\end{algorithm}

\begin{algorithm}[ht]
\caption{Phase 2: Probability-Guided Token Replacement}
\label{alg:phase2}
\begin{algorithmic}[1]
\REQUIRE Edit prompt $c_{\mathrm{edit}}$; cached source reference
\REQUIRE Preservation parameters; pruning ratio $K$
\ENSURE Edited tokens $\{x^{(s)}\}_{s=1}^{S}$
\FOR{$s = 1$ to $S$}
    \STATE $h_{\mathrm{out}}^{(s)}
    \leftarrow\mathrm{PrunedForward}(s,K,c_{\mathrm{edit}})$
    \STATE Compute $p^{(s)}_{\mathrm{edit}}$ from its logits
    \IF{$s < S_{\mathrm{stop}}$}
        \STATE $\{\gamma^{(s)}_i\}\leftarrow\mathrm{SpatialModulation}(A^{(s)},s)$
        \FOR{each token $i$}
            \STATE $b^{(s)}_i \leftarrow \max(\gamma^{(s)}_i - \hat{p}^{(s)}_{\mathrm{src}}(\hat{x}^{(s)}_i), 0)$
            \STATE $x^{*(s)}_i \leftarrow \arg\max_j p^{(s)}_{\mathrm{edit}}(j)$
            \STATE $q_i\leftarrow p^{(s)}_{\mathrm{edit}}(\hat{x}^{(s)}_i)+b^{(s)}_i$
            \IF{$q_i \geq p^{(s)}_{\mathrm{edit}}(x^{*(s)}_i)$}
                \STATE $x^{(s)}_i \leftarrow \hat{x}^{(s)}_i$ \COMMENT{preserve}
            \ELSE
                \STATE $x^{(s)}_i \leftarrow x^{*(s)}_i$ \COMMENT{replace}
            \ENDIF
        \ENDFOR
    \ELSE
        \STATE $x^{(s)}_i \leftarrow \arg\max_j p^{(s)}_{\mathrm{edit}}(j)\ \forall\, i$
    \ENDIF
\ENDFOR
\STATE Edited video $\leftarrow \mathrm{Decode}(\{x^{(s)}\})$
\end{algorithmic}
\end{algorithm}

\begin{algorithm}[ht]
\caption{Residual-Guided Token Pruning}
\label{alg:pruning}
\begin{algorithmic}[1]
\REQUIRE Scale $s$; input states $h_{\mathrm{in}}^{(s)}$
\REQUIRE Previous-scale residual map $r^{(s-1)}$; ratio $K$
\REQUIRE Cross-scale KV cache; edit condition $c_{\mathrm{edit}}$
\ENSURE Output states $h_{\mathrm{out}}^{(s)}$
\IF{$s \notin \{S-1,S\}$}
    \STATE $h_{\mathrm{out}}^{(s)}
    \leftarrow \mathrm{Transformer}(h_{\mathrm{in}}^{(s)})$
\ELSE
    \STATE $\tilde r^{(s)}
    \leftarrow \mathrm{Interpolate}(r^{(s-1)})$
    \STATE $\mathcal{I}_s
    \leftarrow \mathrm{TopKIndices}(\tilde r^{(s)},K)$
    \STATE $h_{\mathrm{out}}^{(s)}
    \leftarrow h_{\mathrm{in}}^{(s)}$ \COMMENT{bypass others}
    \FOR{each Transformer block $\ell$}
        \STATE Update only $h_{\mathrm{out},\mathcal{I}_s}^{(s)}$
        \STATE Use cached cross-scale KV and $c_{\mathrm{edit}}$
    \ENDFOR
\ENDIF
\STATE Forward $h_{\mathrm{out}}^{(s)}$ to the logits head
\end{algorithmic}
\end{algorithm}

\section{Implementation Details}
\label{sec:supp_implementation}

We build Edit-VAR on the 8B-parameter InfinityStar model and generate 81-frame videos at 480p resolution using BF16 inference on a single NVIDIA A800 GPU. Experiments are conducted on Ubuntu 22.04.5 LTS with Python 3.11.14, PyTorch 2.7.1, and CUDA 12.8. We use the same fixed configuration for all 160 editing cases. Table~\ref{tab:implementation_config} summarizes the parameters introduced by our method; all remaining backbone settings follow the official InfinityStar configuration.

\begin{table}[ht]
\centering
{\small
\setlength{\tabcolsep}{5pt}
\begin{tabular}{lc}
\toprule
Parameter & Value \\
\midrule
$\gamma_{\mathrm{start}}$ & 2.00 \\
$\gamma_{\mathrm{end}}^{\mathrm{fg}}$ / $\gamma_{\mathrm{end}}^{\mathrm{bg}}$ & 1.60 / 1.78 \\
Image / video transition scale & 5 / 4 \\
Scale-envelope width $w_t$ & 0.06 \\
Attention center $c_a$ / width $w_a$ & 0.50 / 0.10 \\
Directly extracted attention layers & First 5 of 36 \\
Max direct-attn sequence length & 1200 \\
Scale-decoupling boundary $S_{\mathrm{stop}}$ & 25 \\
Pruned scales / retained ratio $K$ & Final two / 50\% \\
Editing seed & 41 \\
\bottomrule
\end{tabular}
}
\caption{Final implementation configuration. The same settings are used for all editing categories.}
\label{tab:implementation_config}
\end{table}

\paragraph{Attention Extraction and Aggregation.}
We select the attention anchor using a deterministic prompt-differencing algorithm. The algorithm first segments the source and edit prompts into words and aligns the two word sequences to obtain their differing spans. For replacement, attribute, and background edits, it returns the source-side word or phrase in the differing span. For object addition, the differing span contains only the newly inserted edit-side phrase and therefore has no source-side counterpart; the algorithm instead returns the nearest unchanged source-side content word adjacent to the insertion span. For example, when \emph{a soldier} is inserted into a desert scene, the automatically selected source-side anchor is \emph{desert}. No anchor is selected or corrected manually, and no additional language model is used. During source-reference construction, we cache cross-attention from the first repetition of the first five Transformer layers and use only the source-conditioned branch. For a multi-word anchor, attention values over all corresponding text tokens are summed. We then average the resulting maps over attention heads and the selected layers and normalize each scale's map to $[0,1]$. Attention is extracted directly for sequence lengths up to 1200. At higher-resolution scales, we interpolate the map from the nearest available scale using bilinear interpolation for image scales and trilinear interpolation for video scales.

\begin{figure*}[!t]
\centering
\includegraphics[width=\textwidth]{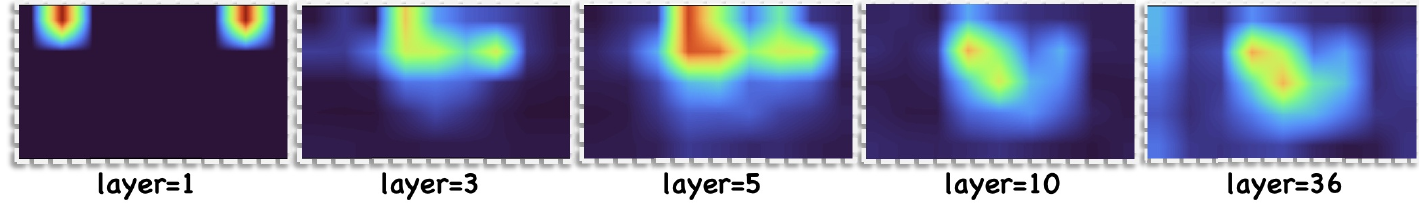}
\caption{Effect of the number of attention layers. We compare localization maps obtained by aggregating the first 1, 3, 5, and 10 layers and all layers. Aggregating the first five layers provides a stable and concentrated localization signal, while using additional layers brings limited qualitative change and incurs greater storage cost. We therefore use the first five layers throughout our experiments.}
\label{fig:supp_attention_layers}
\end{figure*}

\section{Source-Video Reconstruction}
\label{sec:supp_reconstruction}

\begin{figure*}[t]
\centering
 \includegraphics[width=\textwidth]{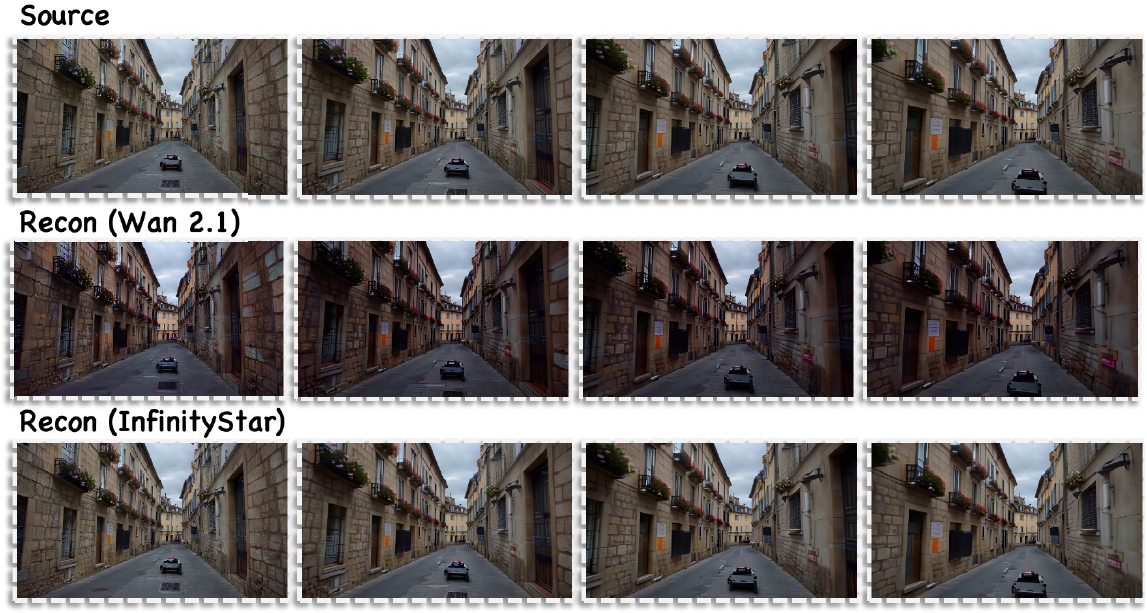}
\caption{Qualitative source-video reconstruction comparison. We show representative source frames together with reconstructions from the inversion-based Wan2.1 pipeline and the direct token-based InfinityStar pipeline, highlighting differences in structural preservation, appearance fidelity, and fine visual details.}
\label{fig:supp_reconstruction}
\end{figure*}

We compare the source-video reconstruction fidelity of pipelines built on the Wan2.1 and InfinityStar backbones. For Wan2.1, each source video is first encoded into the VAE latent space, reconstructed through the inversion-and-regeneration process, and finally decoded back to video space. In contrast, InfinityStar encodes the source video into its complete multi-scale discrete token representation and directly decodes these tokens, without iterative trajectory inversion. We use identical input frames and temporal alignment for both pipelines and report full-frame PSNR, SSIM, and LPIPS. Table~\ref{tab:supp_reconstruction} presents the quantitative comparison, while Figure~\ref{fig:supp_reconstruction} shows representative reconstructions.

\begin{table}[t]
\centering
{\small
\setlength{\tabcolsep}{6pt}
\begin{tabular}{lccc}
\toprule
Pipeline & PSNR$\uparrow$ & SSIM$\uparrow$ & LPIPS$\downarrow$ \\
\midrule
Wan2.1               & 25.62 & 0.898 & 0.124 \\
InfinityStar         & 29.93 & 0.920 & 0.061 \\
\bottomrule
\end{tabular}
}
\caption{Source-video reconstruction fidelity of the Wan2.1 and InfinityStar pipelines. Wan2.1 uses VAE encoding followed by inversion-based reconstruction and decoding, whereas InfinityStar directly reconstructs the video from its encoded multi-scale tokens.}
\label{tab:supp_reconstruction}
\end{table}

\section{Hyperparameter Sensitivity}
\label{sec:supp_sensitivity}

We analyze the sensitivity of our method to three key editing hyperparameters: $\gamma_{\mathrm{end}}^{\mathrm{fg}}$, $\gamma_{\mathrm{end}}^{\mathrm{bg}}$, and $S_{\mathrm{stop}}$. We vary one parameter at a time while fixing all remaining parameters to their default values. The pruning ratio $K$ is studied separately in Section~\ref{sec:supp_efficiency}, where both quality and runtime are considered.

\paragraph{Notation and Parameters.}
For each image or video tower, $N_{\mathrm{tower}}$ denotes its number of scales and $s_{\mathrm{local}}\in\{0,\ldots,N_{\mathrm{tower}}-1\}$ denotes the corresponding local scale index. We normalize the scale position as $t=s_{\mathrm{local}}/(N_{\mathrm{tower}}-1)$. In the scale envelope, $t_c$ is the transition center and $w_t$ controls its width. In the token-wise attention mapping, $c_a$ is the attention transition center and $w_a$ controls its width. The endpoint parameters $\gamma_{\mathrm{end}}^{\mathrm{fg}}$ and $\gamma_{\mathrm{end}}^{\mathrm{bg}}$ determine the late-scale preservation strengths for edit-relevant and preserved regions, respectively. For Scale-Decoupled Generation, $S_{\mathrm{stop}}$ denotes the decoupling boundary: source probabilities and attention maps are cached only for $s<S_{\mathrm{stop}}$, and generation is unconstrained for $s\geq S_{\mathrm{stop}}$. For residual-guided pruning, $K$ is the percentage of tokens retained for Transformer computation. Pruning is applied only to the final two scales, with $K=50\%$ used by default.

\begin{table}[t]
\centering
{\small
\setlength{\tabcolsep}{2.5pt}
\resizebox{\columnwidth}{!}{%
\begin{tabular}{l c cccc}
\toprule
Parameter & Value & CLIP-S$_{\mathrm{edit}}\uparrow$ & PSNR$\uparrow$ & LPIPS$\downarrow$ & Smooth.$\uparrow$ \\
\midrule
\multirow{6}{*}{$\gamma_{\mathrm{end}}^{\mathrm{fg}}$}
& 1.40          & 0.980 & 19.75 & 0.189 & 0.985 \\
& 1.50          & 0.977 & 21.06 & 0.183 & 0.984 \\
& 1.55          & 0.981 & 22.37 & 0.178 & 0.983 \\
& 1.60$^\dagger$ & 0.980 & 22.89 & 0.172 & 0.989 \\
& 1.65          & 0.971 & 22.97 & 0.175& 0.981 \\
& 1.70          & 0.954 & 23.01 & 0.177 & 0.980 \\
\midrule
\multirow{6}{*}{$\gamma_{\mathrm{end}}^{\mathrm{bg}}$}
& 1.60          & 0.982 & 15.41 & 0.289 & 0.980 \\
& 1.70          & 0.973 & 19.77 & 0.201 & 0.988 \\
& 1.74          & 0.980 & 22.78 & 0.180 & 0.984 \\
& 1.78$^\dagger$ & 0.980 & 22.89 & 0.172 & 0.989 \\
& 1.82          & 0.975 & 23.07 & 0.170 & 0.979 \\
& 1.90          & 0.941 & 23.91 & 0.163 & 0.977 \\
\midrule
\multirow{6}{*}{$S_{\mathrm{stop}}$}
& 23          & 0.982 & 17.41 & 0.247 & 0.990 \\
& 24          & 0.982 & 19.33 & 0.225 & 0.986 \\
& 25$^\dagger$ & 0.980 & 22.89 & 0.172 & 0.989 \\
& 26          & 0.979 & 21.73 & 0.184 & 0.981 \\
& 27          & 0.975 & 20.72 & 0.204 & 0.973 \\
& 29          & 0.960 & 18.93 & 0.231 & 0.921 \\
\bottomrule
\end{tabular}
}
}
\caption{Hyperparameter sensitivity analysis. We vary one parameter at a time while keeping the others fixed. $^\dagger$ denotes the default configuration.}
\label{tab:sensitivity}
\end{table}

\paragraph{Foreground Preservation Strength.}
The results exhibit a smooth trade-off as $\gamma_{\mathrm{end}}^{\mathrm{fg}}$ changes. A lower value weakens the source constraint in edit-relevant regions and generally improves target alignment, but may sacrifice local source fidelity and temporal stability. Conversely, a higher value favors source-token preservation and can make the intended modification less complete. The default value of 1.60 lies near the balance between these objectives, and performance remains stable within its neighborhood.

\paragraph{Background Preservation Strength.}
Increasing $\gamma_{\mathrm{end}}^{\mathrm{bg}}$ strengthens source preservation outside the edit-relevant region, improving non-edit fidelity. Excessively strong preservation, however, can constrain tokens near the edit boundary and slightly reduce edit alignment. The results support using distinct foreground and background preservation strengths rather than a spatially uniform control value, with the default 1.78 providing a favorable balance.

\paragraph{Scale-Decoupling Boundary.}
The decoupling boundary presents an intermediate optimum. Applying Scale-Decoupled Generation too early weakens the preservation of source structure, whereas delaying it to too many high-resolution scales over-constrains motion- and pose-dependent details and can introduce texture fragmentation. Setting $S_{\mathrm{stop}}=25$ decouples the source reference after the main layout has been established while leaving sufficient later scales to regenerate coherent edited details.

Overall, the method is stable over a reasonably broad neighborhood of the default configuration. Each parameter produces a predictable quality--preservation trade-off, and the selected configuration lies near the trade-off knee rather than being optimized for a single metric. We use this fixed configuration for all editing cases.

\section{Evaluation Protocol}
\label{sec:supp_evaluation}

\paragraph{Evaluation Set Construction.}
We collect 160 source videos from publicly accessible online sources, with 40 cases for each of object replacement, object addition, background replacement, and attribute editing. We select videos with a clearly visible subject, observable motion, sufficient visual quality for assessing source preservation, and content that supports an unambiguous text-guided edit. This collection is used only as an evaluation set and is not introduced as a new benchmark or dataset contribution.

Let $I_t$, $\hat I_t$, and $M_t$ denote the source frame, edited frame, and edit-region mask at time $t$, respectively. For each editing case, we construct $M_t$ once using SAM2 and use the same mask to evaluate every method; the mask therefore remains independent of the quality or spatial extent of an individual method's output. For object addition, $M_t$ denotes the intended insertion region. The masks are propagated across frames by SAM2.

We compute full-frame CLIP-S between $\hat I_t$ and the complete target prompt. Edit-region CLIP-S is computed from the region specified by $M_t$ and the edit-relevant target phrase, thereby focusing the score on the intended semantic modification rather than unrelated prompt content. To evaluate source-content preservation, PSNR, SSIM, and LPIPS are computed between corresponding source and edited frames over the complementary non-edit region $1-M_t$. The same region-aware implementation is used for every method. Motion Smoothness and Aesthetic Quality are computed using the official VBench implementation; the former uses frame interpolation to measure motion continuity, whereas the latter averages LAION aesthetic predictions over frames.

For every frame-level metric, we first average over the $T_n$ frames of video $n$ and then report the macro-average over the $N$ editing cases:
\begin{equation}
S = \frac{1}{N}\sum_{n=1}^{N}\left(\frac{1}{T_n}\sum_{t=1}^{T_n}s_{n,t}\right).
\end{equation}
Before evaluation, all outputs are temporally sampled to a common frame count. All methods then use identical frames, masks, and metric implementations. Each method is run twice for every editing case under the same fixed configuration, and the reported metric values are averaged over the two runs before aggregation across cases.

\section{Baseline Implementation Details}
\label{sec:supp_baselines}

We evaluate all baselines using their official implementations and author-recommended default configurations, without case-specific parameter tuning. All methods, including ours, are evaluated at a unified 480p output resolution. We only adapt temporal sampling and input formatting when required by each method. Source and target prompts are kept semantically identical across methods, with only method-required formatting changes. VACE-V2V uses the official Wan-based VACE implementation; RAVE and FADE use their official inversion-based pipelines; Wan-Edit uses the official implementation released with FiVE-Bench; and FlowDirector uses its official Wan2.1-based implementation. The corresponding official repositories are \url{https://github.com/ali-vilab/VACE}, \url{https://github.com/RehgLab/RAVE}, \url{https://github.com/EternalEvan/FADE}, \url{https://github.com/MinghanLi/FiVE-Bench}, and \url{https://github.com/Westlake-AGI-Lab/FlowDirector}, respectively.

\section{Additional Comparisons with Baselines}
\label{sec:supp_additional_baselines}

\begin{figure*}[!t]
\centering
\includegraphics[width=\textwidth]{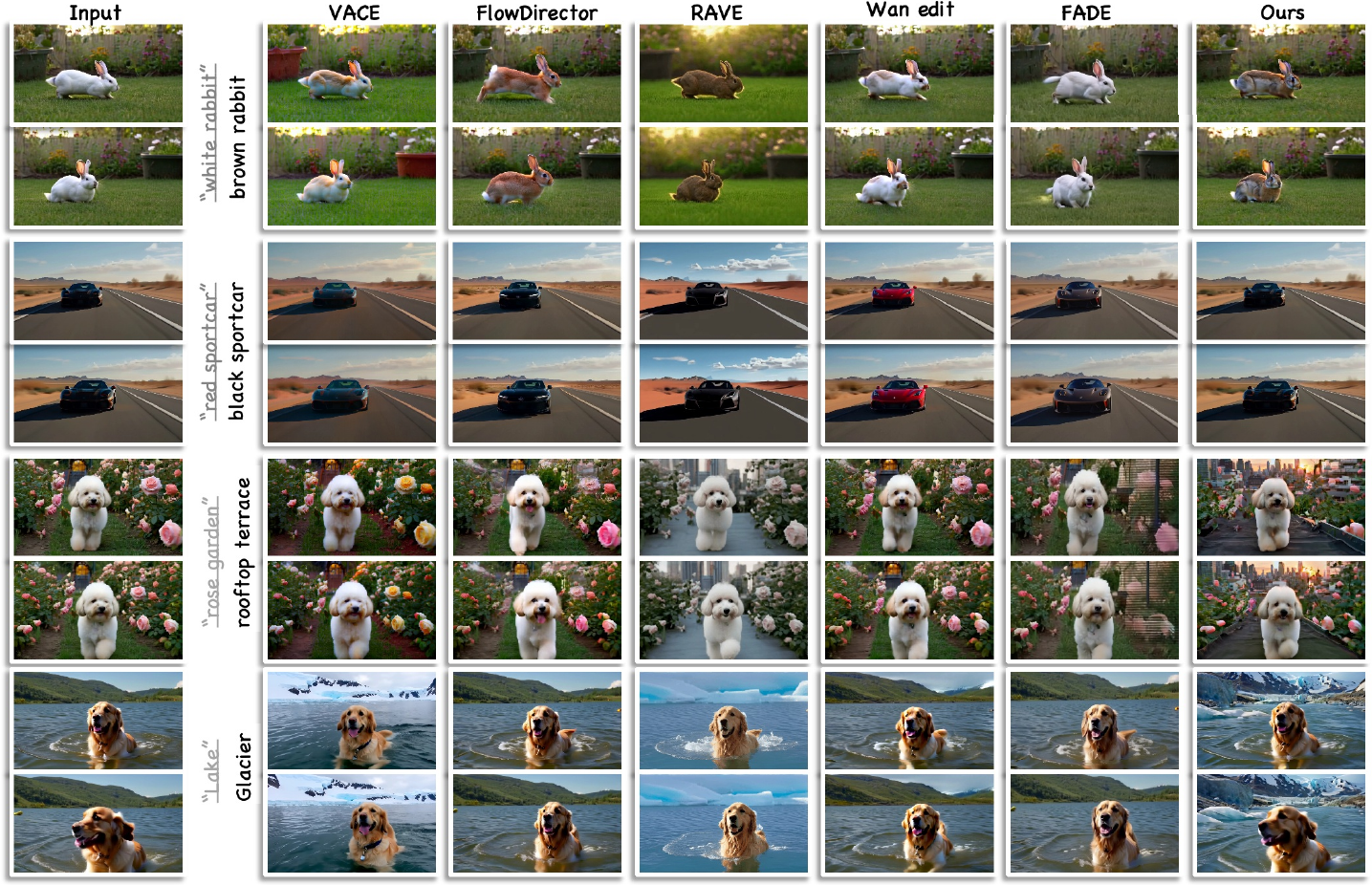}
\caption{Additional qualitative comparisons with the baselines evaluated in the main paper. All methods use the same source video and target editing prompt for each case, with matched temporal sampling and output resolution.}
\label{fig:supp_additional_baseline_comparisons}
\end{figure*}

We provide additional qualitative comparisons with the baselines evaluated in the main paper. These examples complement the main-paper comparison by covering more source videos and editing prompts under the same evaluation protocol. For every case, all methods use the same source video and semantically identical target prompt, and the results are shown with matched temporal sampling and output resolution. Figure~\ref{fig:supp_additional_baseline_comparisons} provides the additional comparisons.

\section{User Study Protocol}
\label{sec:supp_user_study}

We conduct a blind user study with 15 participants on 50 cases randomly selected from the evaluation set. Participants are shown the source video, the target editing prompt, and an anonymized edited result. Method identities are hidden, and the presentation order is randomized. All videos are displayed at the same size and frame rate and can be replayed before a rating is submitted. The same protocol is used for our method and all five baselines.

\begin{figure*}[!t]
\centering
\includegraphics[width=\textwidth]{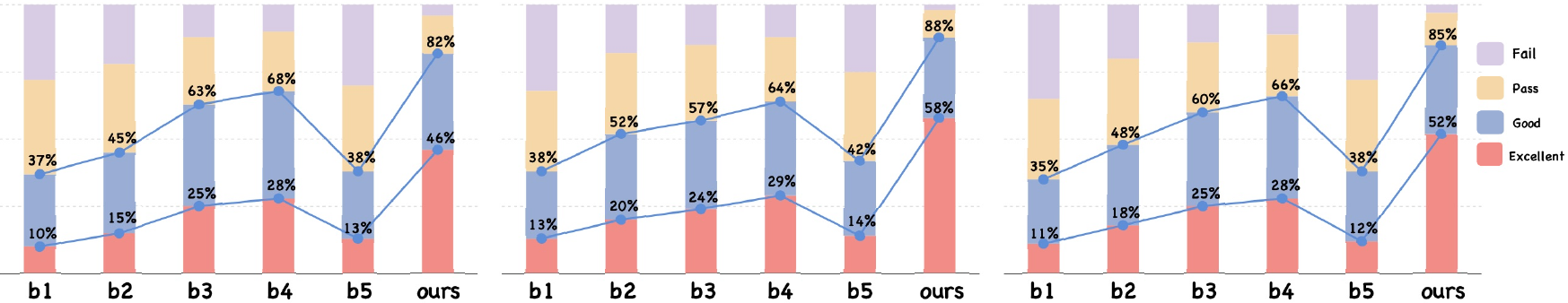}
\caption{User study results. Distributions of the four ordinal ratings (\emph{Fail}, \emph{Pass}, \emph{Good}, and \emph{Excellent}) assigned independently to each method. From left to right, the three panels report editing quality, background preservation, and temporal consistency. }
\label{fig:supp_user_study_results}
\end{figure*}

Participants rate each result independently according to three criteria: (1) \emph{editing quality}, measuring whether the requested modification is correctly and naturally realized; (2) \emph{background preservation}, measuring the fidelity of content outside the intended edit region to the source video; and (3) \emph{temporal consistency}, measuring appearance stability and the absence of flickering or motion artifacts across frames. For each criterion, participants assign one of four ordinal ratings: \emph{Fail}, \emph{Pass}, \emph{Good}, or \emph{Excellent}. Figure~\ref{fig:supp_user_study_results} reports the rating distribution for each method and criterion.

\FloatBarrier

\section{Detailed Efficiency Analysis}
\label{sec:supp_efficiency}

We provide a detailed quality--efficiency analysis of residual-guided token pruning at the final two generation scales. The unoptimized pipeline requires 65.0\,s for each phase. Scale-Decoupled Generation reduces Phase~1 to 19.8\,s ($3.29\times$), giving an unpruned two-phase runtime of 84.8\,s. Here, $K$ denotes the percentage of tokens retained in Phase~2, and $K=50\%$ is our default setting. Random selection at matched token budgets isolates the contribution of residual-based selection from that of reducing the token count alone.

\begin{table*}[!t]
\centering
{\small
\setlength{\tabcolsep}{4pt}
\resizebox{\textwidth}{!}{%
\begin{tabular}{l c cc cc ccccc}
\toprule
\multirow{2}{*}{Selection} & \multirow{2}{*}{Kept $K$} & \multicolumn{2}{c}{Phase~2} & \multicolumn{2}{c}{Two-Phase Total} & \multicolumn{5}{c}{Quality} \\
\cmidrule(lr){3-4} \cmidrule(lr){5-6} \cmidrule(lr){7-11}
& & Time (s)$\downarrow$ & Speedup$\uparrow$ & Time (s)$\downarrow$ & Speedup$\uparrow$ & CLIP-S$_{\mathrm{full}}\uparrow$ & CLIP-S$_{\mathrm{edit}}\uparrow$ & Smooth.$\uparrow$ & Aesth.$\uparrow$ & LPIPS$\downarrow$ \\
\midrule
No pruning       & 100\% & 65.0 & 1.00$\times$ & 84.8 & 1.00$\times$ & .979 & .980 & .989 & .611 & .172 \\
\midrule
Residual-guided  & 90\%  & 61.2 & 1.06$\times$ & 81.0 & 1.05$\times$ & .979 & .978 & .989 & .610 & .175 \\
Residual-guided  & 70\%  & 51.2 & 1.27$\times$ & 70.9 & 1.19$\times$ & .975 & .977 & .989 & .608 & .175 \\
Residual-guided  & 50\%  & 43.9 & 1.48$\times$ & 63.7 & 1.33$\times$ & .972 & .976 & .988 & .609 & .176 \\
Residual-guided  & 30\%  & 39.4 & 1.65$\times$ & 59.2 & 1.43$\times$ & .965 & .970 & .975 & .593 & .204 \\
\midrule
Random           & 70\%  & 51.2 & 1.27$\times$ & 70.9 & 1.19$\times$ & .956 & .961 & .963 & .592 & .211 \\
Random           & 50\%  & 43.9 & 1.48$\times$ & 63.7 & 1.33$\times$ & .943 & .954 & .976 & .589 & .235 \\
\bottomrule
\end{tabular}
}
}
\caption{Quality--efficiency trade-off of residual-guided token pruning. Pruning is applied to the final two generation scales. Runtime is measured on a single NVIDIA A800 GPU with 81 frames. Phase~2 speedup is relative to its 65.0\,s unpruned runtime; two-phase speedup is relative to the 84.8\,s pipeline with Scale-Decoupled Generation and no pruning.}
\label{tab:supp_pruning_efficiency}
\end{table*}

\section{Per-Edit-Type Results}
\label{sec:supp_per_type}
Table~\ref{tab:per_type} reports per-edit-type quantitative results to complement the aggregated comparison in the main paper.

\begin{table*}[!t]
\centering
{\small
\setlength{\tabcolsep}{6pt}
\resizebox{\textwidth}{!}{%
\begin{tabular}{ll cc cc ccc}
\toprule
\multirow{2}{*}{Edit Type} & \multirow{2}{*}{Method} & \multicolumn{2}{c}{Target Alignment} & \multicolumn{2}{c}{Video Quality} & \multicolumn{3}{c}{Non-edit Preservation} \\
\cmidrule(lr){3-4} \cmidrule(lr){5-6} \cmidrule(lr){7-9}
& & CLIP-S$_{\mathrm{full}}\uparrow$ & CLIP-S$_{\mathrm{edit}}\uparrow$ & Smooth.$\uparrow$ & Aesth.$\uparrow$ & PSNR$\uparrow$ & SSIM$\uparrow$ & LPIPS$\downarrow$ \\
\midrule
\multirow{2}{*}{Object Replacement}
& Best Baseline & 0.941 & 0.957 & 0.988 & 0.611 & 20.76 & 0.801 & 0.190 \\
& Ours          & 0.957 & 0.970 & 0.991 & 0.613 & 21.87 & 0.815 & 0.176 \\
\midrule
\multirow{2}{*}{Object Addition}
& Best Baseline & 0.976 & 0.984 & 0.970 & 0.581 & 20.07 & 0.788 & 0.199 \\
& Ours          & 0.979 & 0.985 & 0.988 & 0.603 & 22.59 & 0.823 & 0.174 \\
\midrule
\multirow{2}{*}{Background Replacement}
& Best Baseline & 0.956 & 0.963 & 0.977 & 0.580 & 19.73 & 0.781 & 0.209  \\
& Ours          & 0.982 & 0.982 & 0.988 & 0.611 & 23.87 & 0.828 & 0.169\\
\midrule
\multirow{2}{*}{Attribute Editing}
& Best Baseline & 0.965 & 0.959 & 0.977 & 0.582 & 18.92 & 0.791 & 0.217 \\
& Ours          & 0.970 & 0.967 & 0.985 & 0.609 & 21.99 & 0.818 & 0.185 \\
\bottomrule
\end{tabular}
}
}
\caption{Per-edit-type quantitative comparison. Results broken down by editing category.}
\label{tab:per_type}
\end{table*}

\FloatBarrier

\section{Additional Qualitative Results}
\label{sec:supp_qualitative}
\begin{figure*}[!t]
\centering
\includegraphics[width=0.95\textwidth]{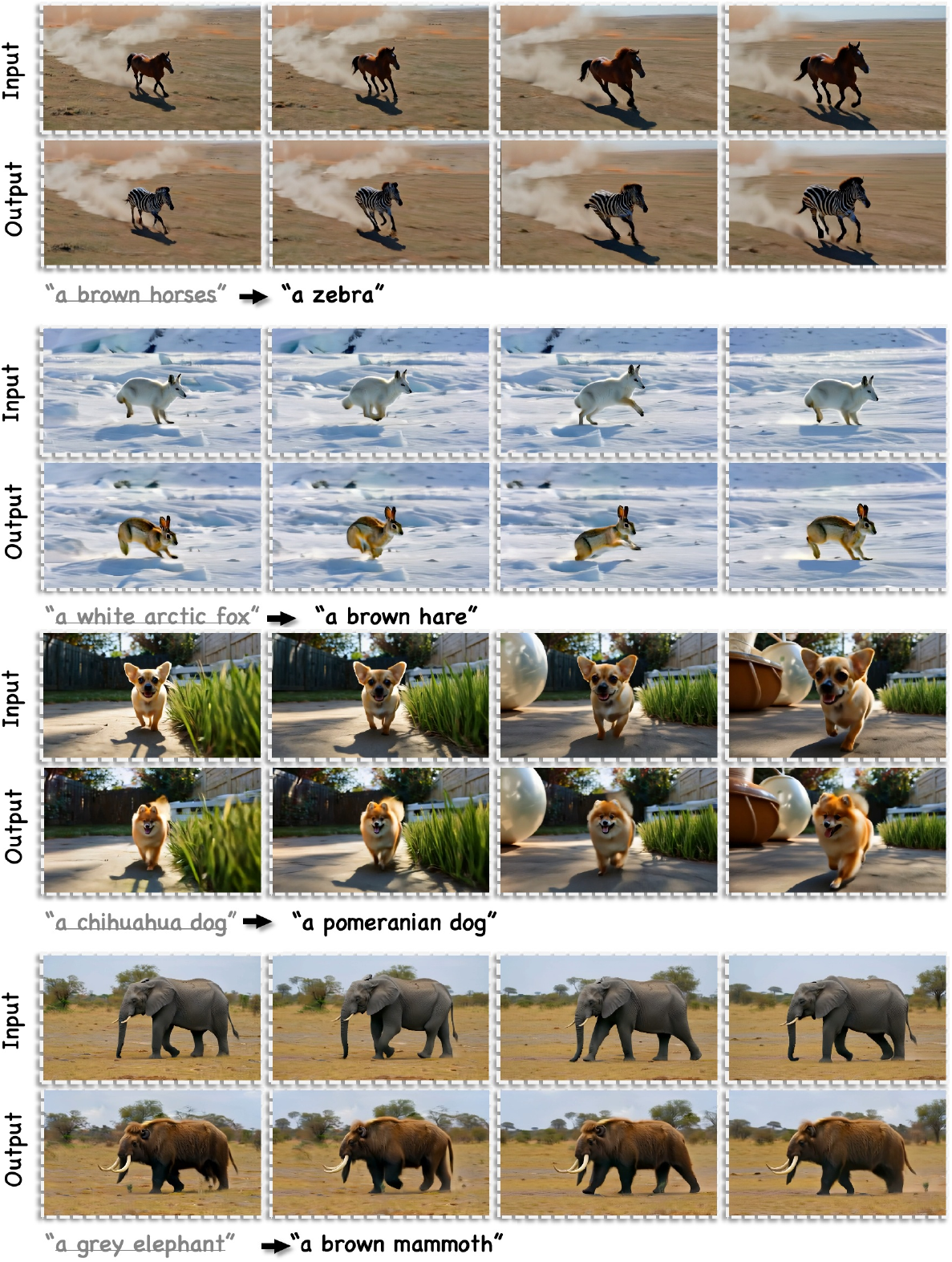}
\caption{Additional results: Object replacement.}
\label{fig:supp_replacement}
\end{figure*}

\begin{figure*}[!t]
\centering
\includegraphics[width=0.95\textwidth]{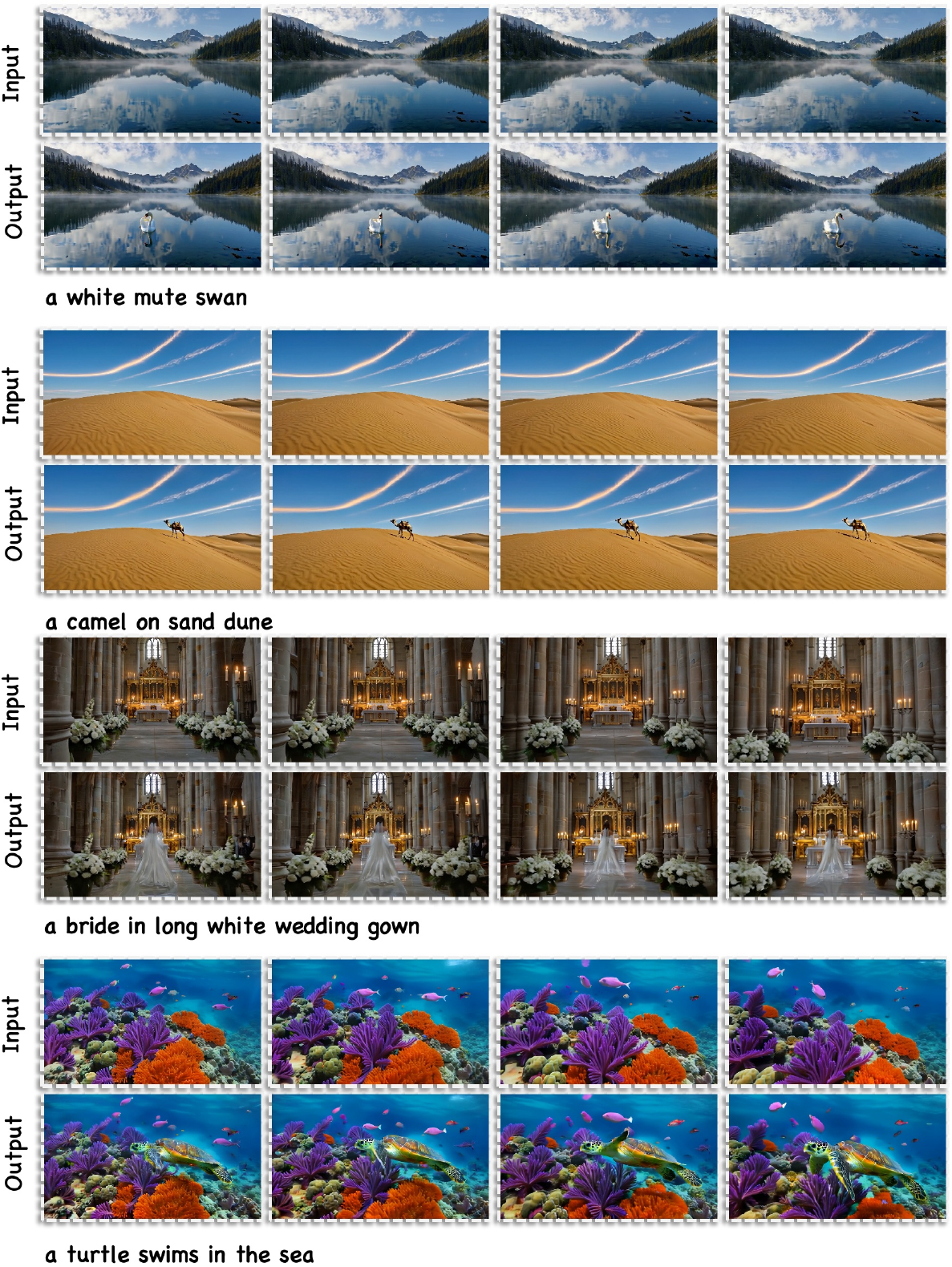}
\caption{Additional results: Object addition.}
\label{fig:supp_addition}
\end{figure*}

\begin{figure*}[!t]
\centering
\includegraphics[width=0.95\textwidth]{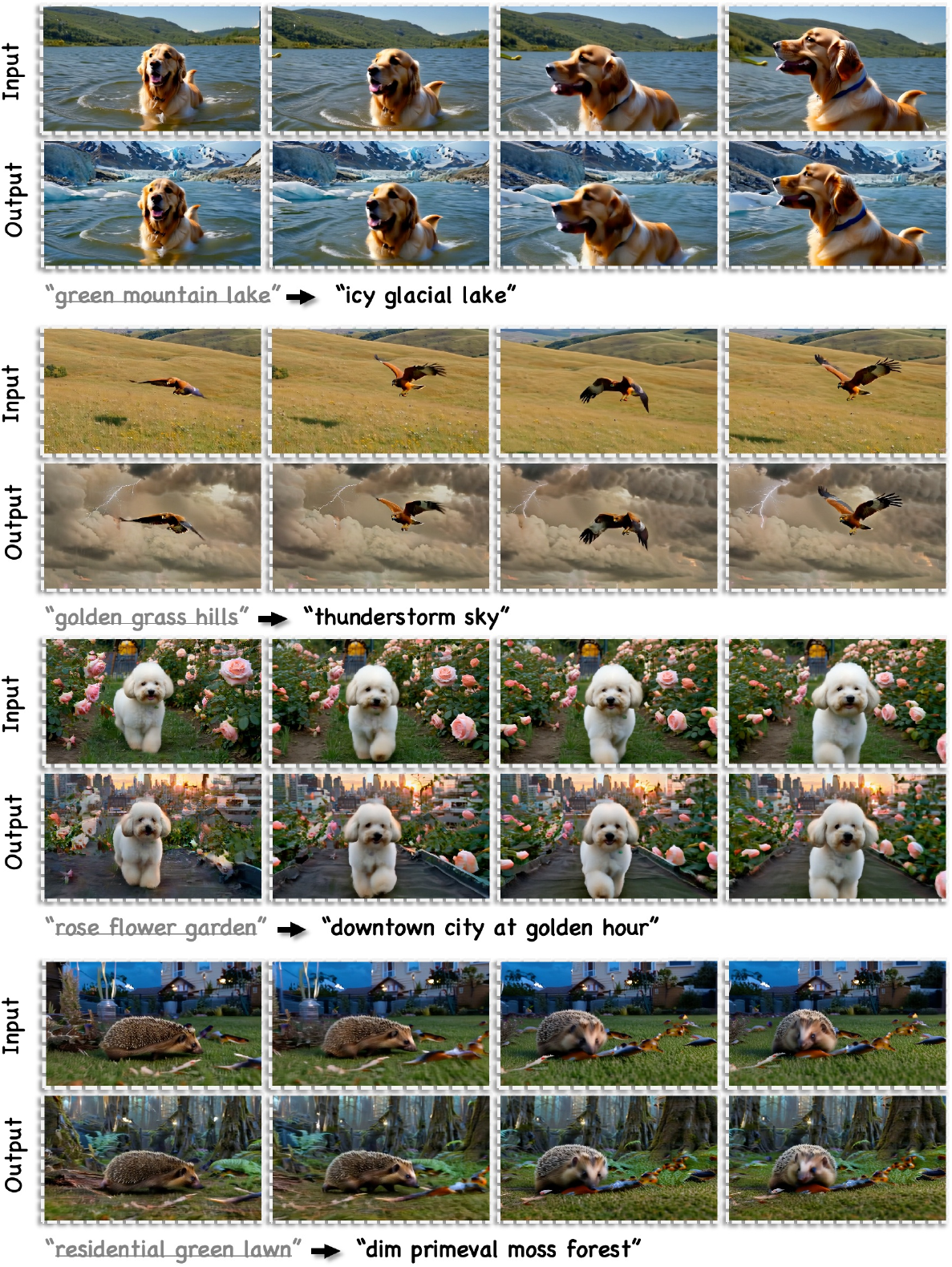}
\caption{Additional results: Background replacement.}
\label{fig:supp_background}
\end{figure*}

\begin{figure*}[!t]
\centering
\includegraphics[width=0.95\textwidth]{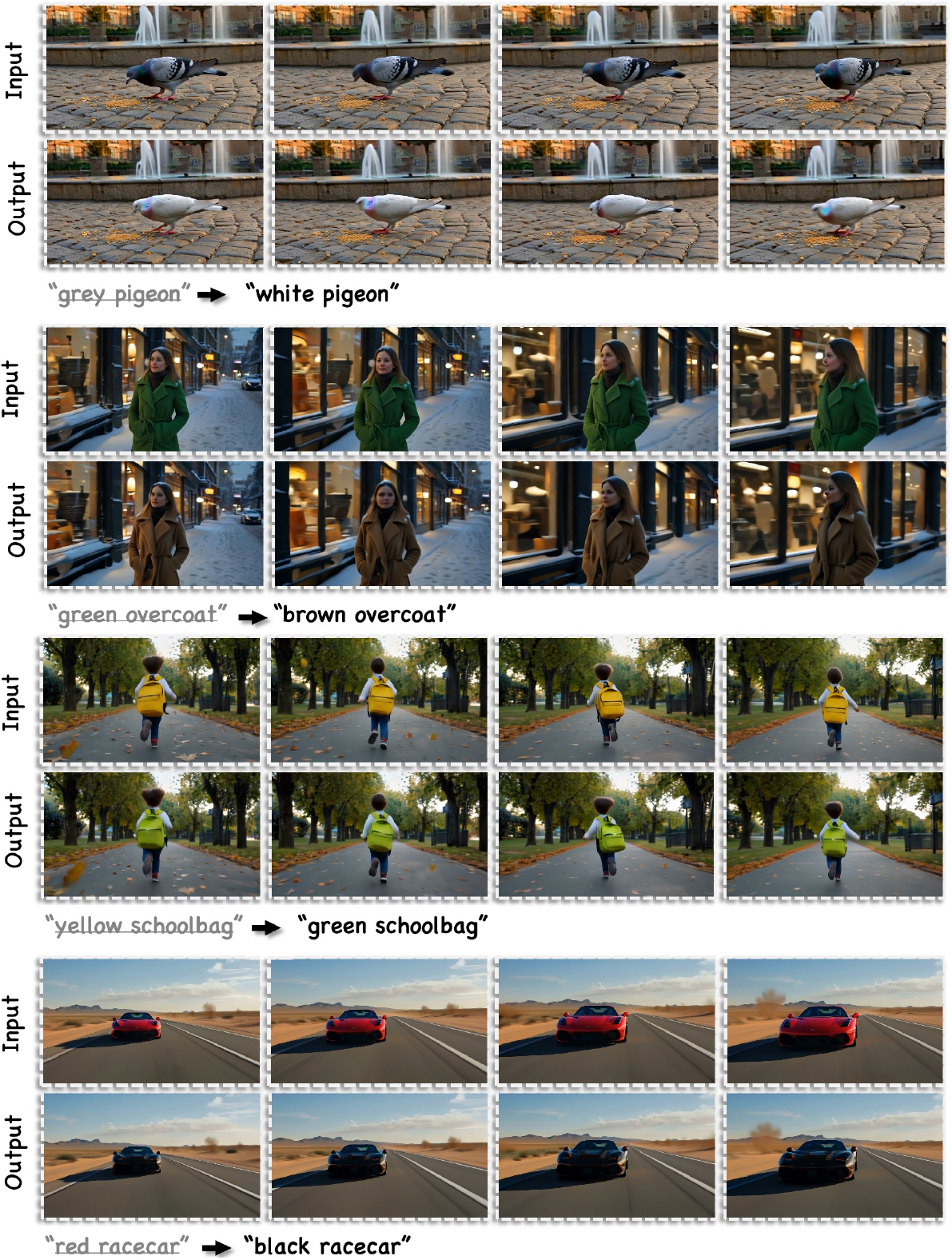}
\caption{Additional results: Attribute editing.}
\label{fig:supp_attribute}
\end{figure*}

Figures~\ref{fig:supp_replacement}, \ref{fig:supp_addition}, \ref{fig:supp_background}, and~\ref{fig:supp_attribute} present additional results for object replacement, object addition, background replacement, and attribute editing, respectively.

\FloatBarrier

\section{Failure Cases}
\label{sec:supp_failure}
We show representative failure cases to discuss the limitations of our method. Our method may struggle when the requested edit requires a substantial structural change from the source video, since the source-preservation constraints favor retaining the original layout, pose, and motion. Failure may also occur when the editing prompt is ambiguous or insufficiently descriptive, in which case the model may misinterpret the intended target or generate incomplete semantic changes. These limitations can lead to residual source structures, implausible local deformations, or edited content that does not fully match the intended concept.

\begin{strip}
\begin{minipage}{\textwidth}
\centering
\includegraphics[width=0.95\linewidth]{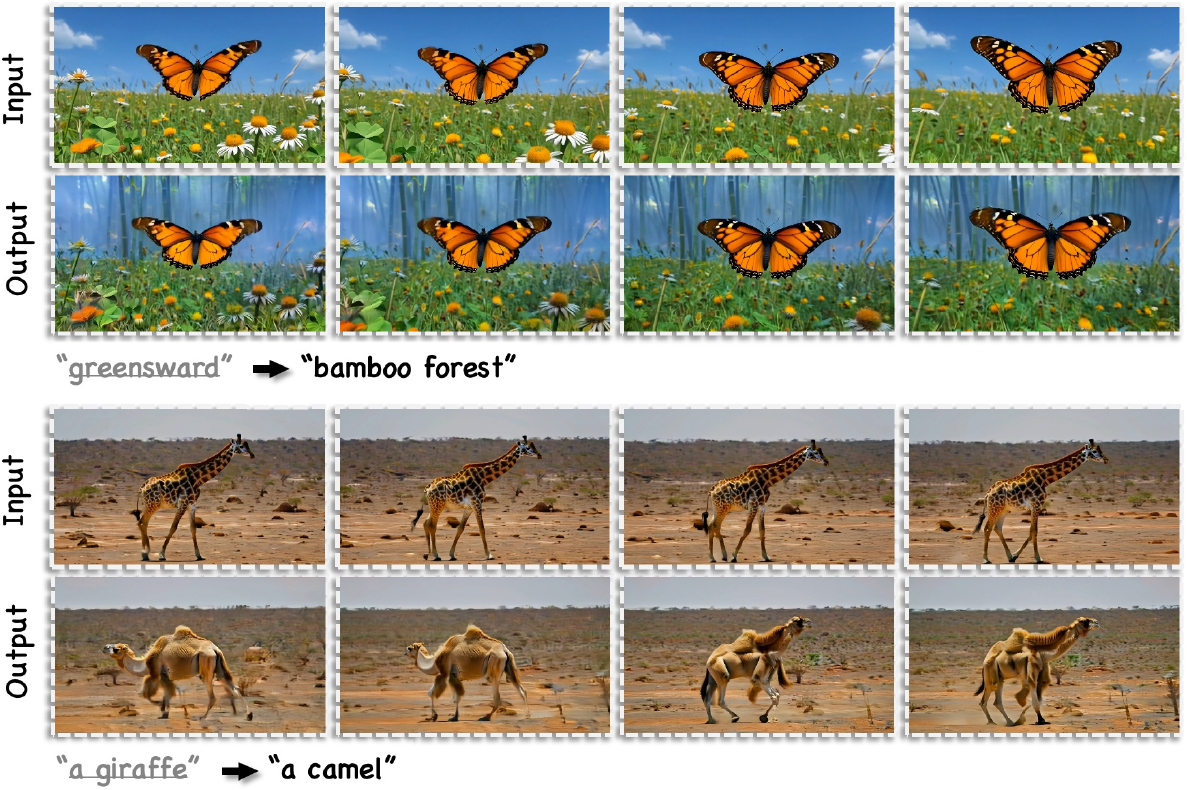}
\captionsetup{hypcap=false}
\captionof{figure}{Failure cases involving large structural changes.}
\captionsetup{hypcap=true}
\label{fig:failure}
\end{minipage}
\end{strip}

\par\noindent\mbox{}\par